\documentclass{article}

     \usepackage[nonatbib, preprint]{neurips_2020}

\usepackage[utf8]{inputenc} 
\usepackage[T1]{fontenc}    
\usepackage{hyperref}       
\usepackage{url}            
\usepackage{booktabs}       
\usepackage{amsfonts}       
\usepackage{nicefrac}       
\usepackage{microtype}      
\usepackage{xcolor}
\usepackage{enumitem}
\usepackage{graphicx}

\usepackage{caption}

\title{
Short-Term Solar Irradiance Forecasting Using Calibrated Probabilistic Models
}
\author{%
    Eric Zelikman\thanks{Equal contribution.} ~$^1$, Sharon Zhou\footnotemark[1]~~$^1$, Jeremy Irvin\footnotemark[1]~~$^1$\\ \textbf{Cooper Raterink$^1$, Hao Sheng$^1$,  Anand Avati$^1$} \\ \textbf{Jack Kelly$^2$, Ram Rajagopal$^1$, Andrew Y. Ng\thanks{Equal contribution.} ~$^1$, David Gagne\footnotemark[2]~~$^3$ }\\
    $^1$Stanford University, $^2$Open Climate Fix, $^3$National Center for Atmospheric Research\\
    \small{\texttt{\{ezelikman,sharonz,jirvin16,crat,avati\}}\texttt{@cs.stanford.edu,haosheng@stanford.edu}} \\
    \small{\texttt{jack@openclimatefix.org}},
    \small{\texttt{ramr@stanford.edu}}, 
    \small{\texttt{ang@cs.stanford.edu}}, 
    \small{\texttt{dgagne@ucar.edu}}
}

\begin{document}

\maketitle

\newif\ifshowcomments
\showcommentstrue
\ifshowcomments
\newcommand\jeremy[1]{\textcolor{red}{[jirvin: #1]}}
\newcommand\sharon[1]{\textcolor{blue}{[sharon: #1]}}
\newcommand\eric[1]{\textcolor{green}{[eric: #1]}}
\else
\newcommand\jeremy[1]{}
\newcommand\sharon[1]{}
\newcommand\eric[1]{}
\fi

\vspace{-15px}
\begin{abstract}
Advancing probabilistic solar forecasting methods is essential to supporting the integration of solar energy into the electricity grid. In this work, we develop a variety of state-of-the-art probabilistic models for forecasting solar irradiance. We investigate the use of post-hoc calibration techniques for ensuring well-calibrated probabilistic predictions. We train and evaluate the models using public data from seven stations in the SURFRAD network, and demonstrate that the best model, NGBoost, achieves higher performance at an intra-hourly resolution than the best benchmark solar irradiance forecasting model across all stations. Further, we show that NGBoost with CRUDE post-hoc calibration achieves comparable performance to a numerical weather prediction model on hourly-resolution forecasting.
\end{abstract}

  \vspace{-5px}
  \begin{figure}[h]
  \vspace{-4px}
  \centering
  \includegraphics[width=0.7\textwidth]{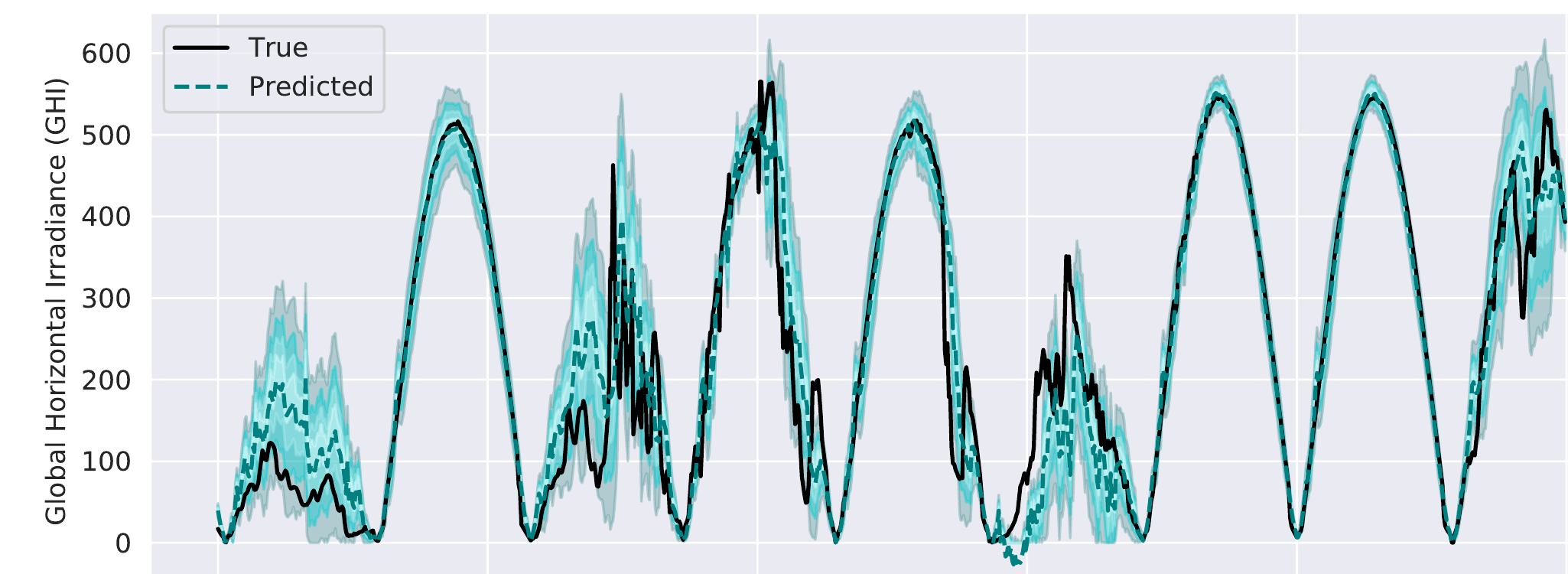}\\
  \vspace{-1px}
  \tiny{Time ($\approx$ 8 days in January 2018)}
  \caption{Several days of 30-min horizon NGBoost irradiance forecasts on the Sioux Falls station~\cite{augustine2000surfrad}.}
  \vspace{-4px}
  \label{pull}
\end{figure}
  \vspace{-11px}
\section{Introduction}
\label{intro}
Increasing adoption of renewable energy in the electricity sector is essential to the reduction of anthropogenic greenhouse gas emissions, but significant production increases must occur in order to phase out fossil fuel use \cite{owusu2016review,elzinga2015energy}. Solar power production has grown dramatically, ranking in the top two electricity-generating capacity additions over the past seven years in the U.S., with 40\% of new electric capacity in the grid coming from solar in 2019~\cite{seia2020solar}. However, due to the high volatility and intermittency of solar energy, solar forecasting methods have become necessary to increase the penetration of solar power into the grid while ensuring power system cost-effectiveness and security \cite{antonanzas2016review}. These methods often forecast solar irradiance, which is widely available and correlates strongly with solar photovoltaic (PV) output \cite{raza2016recent}. While solar forecasting has received extensive attention in the literature \cite{mathiesen2011evaluation,inman2013solar,antonanzas2016review,ahmed2020review}, it was noted by \cite{yang2019guideline} that the overwhelming majority of solar forecasting methods are not probabilistic. Solar forecasting methods that characterize uncertainty have the potential to aid real-time grid integration of solar energy and help gauge when to deploy new storage \cite{haupt2019use,taheri2019energy}. For example, anticipating periods of high uncertainty in solar power production may allow a utility to proactively store energy \cite{haupt2019use} and anticipating periods of low uncertainty can reduce the need for operating reserves, which are often gas-powered \cite{wu2015integrating}.

Recent advancements in probabilistic modeling and uncertainty estimation present an opportunity to substantially improve solar irradiance forecasting. The development of probabilistic solar irradiance forecasting models is advancing \cite{van2018review,li2020review}, but most of these methods use numerical weather prediction (NWP) and traditional statistical approaches for producing probabilistic outputs \cite{alessandrini2015analog,lauret2017probabilistic}. Moreover, many of the traditional approaches like NWP models are primarily useful on hourly timescales and introduce extensive computational overhead, whereas the best choice for short-term, intra-hourly forecasting has remained open \cite{doubleday2020benchmark}. Deep learning methods for probabilistic solar forecasting have recently begun to emerge, but have not been widely adopted as they have not yet demonstrated superior performance over traditional methods \cite{wang2019review}. Furthermore, recent advances in machine learning have developed post-hoc calibration techniques for encouraging well-calibrated predictions \cite{kuleshov2018accurate,crude}, but these methods have yet to be used in solar irradiance forecasting \cite{doubleday2020benchmark}.

In this work, we develop and validate several probabilistic solar irradiance forecasting models using public data from seven meteorological stations in the U.S.~\cite{augustine2000surfrad}. We compare the performance of the probabilistic models together with several modern post-hoc calibration methods to state-of-the-art benchmarks from \cite{doubleday2020benchmark}. We demonstrate that the best model outperforms the alternatives across all seven stations for intra-hourly forecasting and performs comparably to numerical weather prediction (NWP) models on hourly-resolution forecasting. This work advances solar irradiance forecasting by developing state-of-the-art probabilistic models and associated post-hoc calibration techniques for significant improvements in solar forecasting.

\section{Methods}
\label{methods}
\textbf{Data.} We use public data from NOAA's Surface Radiation (SURFRAD) network \cite{augustine2000surfrad}, consisting of seven stations throughout the continental United States that measure a variety of meteorological variables. Relative humidity, wind speed, wind direction, air pressure, time of day, solar zenith angle, air temperature, and the five previous irradiance values up to the forecast time are used as input, and daytime global horizontal irradiance (GHI) as output. We compute the clearness index as the ratio of the measured terrestrial GHI values to extraterrestrial radiation \cite{kato2016prediction}. Extraterrestrial radiation estimates are taken from the publicly available CAMS McClear Sky historical clear sky irradiance estimates at the stations \cite{granier2019copernicus}, as performed in \cite{doubleday2020benchmark}. Using the clearness index allows the model to account for predictable changes in irradiance as a function of deterministic factors and results in a stationary time-series \cite{kato2016prediction}. We use 2016 SURFRAD data to learn model parameters, 2017 data to fit calibration method parameters, and 2018 data to evaluate and compare the models. 

\begin{figure}[b]
  \centering
  \vspace{-10px}
  \includegraphics[width=0.85\textwidth]{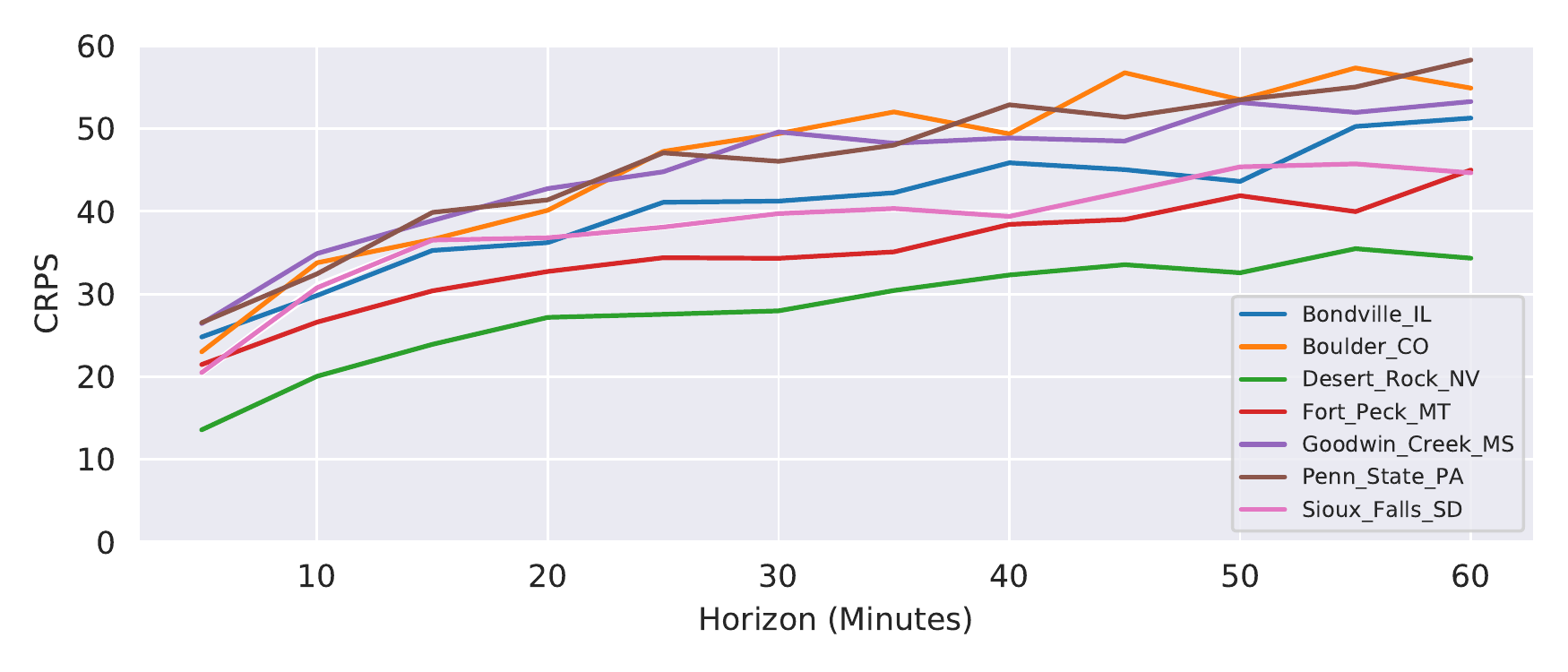}
  \vspace{-10px}
  \caption{CRPS as a function of forecast horizon for the best model (uncalibrated NGBoost).}
  \label{crpscurve}
\end{figure}

\textbf{Models.} We develop four probabilistic models which output a probability distribution over the outcome space instead of a point prediction: a Gaussian process regression model, a neural network with uncertainty based on dropout variation (Dropout Neural Network), a neural network whose predictions parameterize a Gaussian distribution optimized to maximize likelihood (Variational Neural Network), and a decision-tree based model using natural gradient boosting (NGBoost) assuming a Gaussian output distribution \cite{duan2019ngboost}. Hyperparameters are provided in the appendix.

\textbf{Calibration Methods.} We explore the use of post-hoc probabilistic calibration methods for encouraging well-calibrated predictions, as they have long been shown to improve the performance of weather models \cite{stewart1992effects}. Post-hoc calibration methods aim to maximize the calibration score (or reliability) and sharpness (or resolution) of a probabilistic model using a held-out calibration dataset \cite{gneiting2007probabilistic,crude}. The calibration score penalizes overconfident or underconfident models; for example, one can evaluate the uniformity of the observations' percentiles within their predicted probability distributions \cite{gneiting2007probabilistic}. We evaluate three post-hoc probabilistic calibration methods, which we refer to as the Kuleshov method \cite{kuleshov2018accurate}, CRUDE \cite{crude}, and the MLE method inspired by \cite{levi2019evaluating}:
\begin{itemize}[topsep=0pt,leftmargin=0.5cm]
\item The Kuleshov method is inspired by Platt scaling: the method employs isotonic regression on a calibration dataset, such that a calibrated prediction at a requested percentile $p$ finds the model-predicted percentile $\hat{p}$ which was greater than the requested portion $p$ of the calibration data, and returns the model prediction at $\hat{p}$ \cite{kuleshov2018accurate}.

\item CRUDE calculates the $z$-scores of all errors on a calibration dataset $(\hat{\mu}(x) - y) / {\hat{\sigma}}(x)$, where $\hat{\mu(x)}$ is the predicted mean, $y$ is the observation, and $\hat{\sigma}(x)$ is the predicted standard deviation \cite{crude}. Then, to make a prediction at a given percentile, the calibration-set $z$-score at the percentile is calculated and multiplied by the predicted standard deviation. A constant shift to $\hat{\mu}(x)$ is learned to maximize the calibration score on the calibration set \cite{crude}.

\item The MLE method calculates the constant shift and scale of the predicted distributions derived with maximum likelihood estimation, assuming a Gaussian distribution. This is inspired by \cite{levi2019evaluating}, which proposed that a simpler maximum likelihood approach was less prone to overfit.
\end{itemize}

\textbf{Performance Metrics.} All models are trained and calibrated on the clearness index, and evaluated on the irradiance, as in \cite{doubleday2020benchmark}. We primarily evaluate the probabilistic models using the continuous ranked probability score (CRPS), a widely used metric which balances calibration and sharpness \cite{candille2005evaluation}. After being shown to be effective for evaluating probabilistic forecasting in other meteorological contexts \cite{avati2020countdown}, CRPS has been recently proposed as a standard performance metric for probabilistic irradiance forecasting \cite{lauret2019verification,doubleday2020benchmark}. \cite{avati2020countdown} presents an intuitive definition of CRPS as the area between each predicted CDF and a step function at the observed value. This requires calibration and rewards sharpness, while being less sensitive to outliers than MLE \cite{avati2020countdown}. We also record the calibration and sharpness metrics discussed in \cite{kuleshov2018accurate} and \cite{crude}. A visualization of the calibration curves for the four models using various post-hoc calibration procedures is shown in Figure~\ref{calcurves}.

\section{Results}

\captionsetup[table]{skip=3pt}
\begin{table}[]
\centering
\resizebox{1.02\textwidth}{!}{%
\hspace{-15px}
\begin{tabular}{@{}l|rrrr|rrrr|rrrr|rrrr@{}}
\toprule \multicolumn{1}{c|}{\textbf{Station}}
 & \multicolumn{4}{c|}{\textbf{Gaussian Process}} & \multicolumn{4}{c|}{\textbf{Dropout Neural Network}} & \multicolumn{4}{c|}{\textbf{Variational Neural Net}} & \multicolumn{4}{c}{\textbf{NGBoost}} \\ \midrule
\textit{} & \multicolumn{1}{c}{\textit{None}} & \multicolumn{1}{c}{MLE} & \multicolumn{1}{c}{C} & \multicolumn{1}{c|}{Kul.} & \multicolumn{1}{c}{\textit{None}} & \multicolumn{1}{c}{MLE} & \multicolumn{1}{c}{C} & \multicolumn{1}{c|}{Kul.} & \multicolumn{1}{c}{\textit{None}} & \multicolumn{1}{c}{MLE} & \multicolumn{1}{c}{C} & \multicolumn{1}{c|}{Kul.} & \multicolumn{1}{c}{\textit{None}} & \multicolumn{1}{c}{MLE} & \multicolumn{1}{c}{C} & \multicolumn{1}{c}{Kul.} \\
Bondville, IL & 101.3 & 53.2 & 48.5 & 48.6 & 48.5 & 46.0 & 43.6 & 44.0 & 42.0 & 42.0 & 41.8 & 41.9 & \textbf{40.5} & \textbf{40.5} & 40.6 & 40.6 \\
Boulder, CO & 110.9 & 61.7 & 56.4 & 56.5 & 59.3 & 55.8 & 53.3 & 53.9 & 48.6 & 48.9 & 48.3 & 48.6 & \textbf{45.9} & 46.1 & 46.0 & 46.2 \\
Desert Rock, NV & 96.6 & 44.3 & 35.4 & 35.7 & 37.2 & 40.8 & 36.1 & 36.2 & 31.4 & 32.5 & 30.0 & 30.3 & 27.9 & 30.1 & \textbf{27.8} & 28.2 \\
Fort Peck, MT & 97.5 & 50.7 & 43.6 & 43.4 & 41.6 & 41.9 & 38.9 & 39.0 & 37.9 & 46.8 & 37.5 & 37.6 & \textbf{34.8} & 35.2 & 35.0 & 34.9 \\
Goodwin Creek, MS & 119.2 & 59.8 & 54.7 & 54.9 & 57.9 & 53.3 & 51.6 & 51.5 & 46.9 & 46.9 & 46.7 & 46.9 & \textbf{44.8} & 45.0 & \textbf{44.8} & 45.1 \\
Penn State, PA & 111.6 & 58.8 & 53.9 & 53.3 & 56.5 & 51.2 & 49.5 & 48.0 & 47.4 & 47.4 & 47.3 & 47.0 & \textbf{46.0} & 46.6 & 46.1 & \textbf{46.0} \\
Sioux Falls, SD & 107.2 & 54.4 & 49.3 & 49.5 & 48.0 & 46.0 & 43.4 & 43.7 & 43.8 & 41.8 & 42.4 & 43.0 & \textbf{37.9} & 39.1 & 38.0 & 38.4 \\ \bottomrule
\end{tabular}%
}
\caption{\textbf{Per-Station Model-Calibration Performance.} Average test CRPS of the probabilistic models with different calibration methods across all seven stations, averaged over each 5-minute horizon from 5 minutes to an hour. The columns correspond to the calibration methods discussed in Section~\ref{methods},  with Kul. for Kuleshov \cite{kuleshov2018accurate} and C for CRUDE \cite{crude}.}
\label{stations-models}
\vspace{-10px}
\end{table}

For intra-hourly forecasting which evaluates each five minute forecast horizon up to an hour, NGBoost attained the best test CRPS scores across all seven stations. For hourly-resolution forecasting, which evaluates each hourly forecast up to six hours, NGBoost performed comparably to the NWP models. Post-hoc calibration improved the test CRPS scores of all models except for intra-hourly NGBoost. As shown in Figure~\ref{pull}, where each color corresponds to an additional 10\% interval (10th-90th percentiles), NGBoost quickly responds to changes in uncertainty, performing well under high and low uncertainty. Additionally, as highlighted in Figure~\ref{crpscurve}, there is a steady increase in CRPS as the horizon increases, with stations being consistent in terms of their relative difficulty, e.g., Desert Rock is consistently predictable across methods and horizons, and Penn State and Boulder are more difficult. 

\textbf{Intra-hourly.} On the intra-hourly forecasting task, the NGBoost models outperformed the best models in \cite{doubleday2020benchmark} across all stations. The test CRPS scores of all models, with all post-hoc calibration methods and without, are comprehensively reported in Table~\ref{stations-models}. In \cite{doubleday2020benchmark}, the Markov-chain Mixture model (MCM), outperformed several high-quality baselines including smart persistence ensembles and a Gaussian error distribution. The CRPS scores of NGBoost and MCM on each of the stations are shown in Table~\ref{ngboostvs}. Performance improvement of NGBoost over MCM ranged from 5\% to over 15\%.

\textbf{Hourly.} On the hourly-resolution task, we primarily compared NGBoost with various post-hoc calibrations to numerical weather prediction (NWP) models, as done in \cite{doubleday2020benchmark} due to the lower temporal resolution of NWP models. NGBoost(+C) was the best model on three stations, and the European Centre for Medium-Range Weather Forecasts (ECMWF) ensemble model, a standard NWP model for meteorological forecasting, was the best model on another three stations. The ECMWF control Gaussian (GAU) error distribution was the best performing model on the other station \cite{doubleday2020benchmark}. The NGBoost results on the hourly-resolution forecasting task in Table~\ref{ngboostvs} suggest that this solution can reach performance which is comparable to that of NWP models. 

\textbf{Calibration.} Across post-hoc calibration methods, CRUDE and Kuleshov led to more substantial performance improvements overall than the MLE method. Notably, CRUDE and Kuleshov improve the calibration metric of NGBoost at the expense of sharpness, with CRUDE reducing the average calibration error across all stations from $0.040$ to $0.031$ but worsening the mean sharpness on the clearness index predictions from $0.192$ to $0.215$ (lower is better). The calibration of NGBoost may result from several factors: the variational neural network is also well calibrated, but likely slightly overfit - the use of the natural gradient by NGBoost is suggested to reduce overfitting \cite{duan2019ngboost}; additionally, the under-confidence of the Gaussian process suggests that investigation of alternative priors may be warranted \cite{schulz2018tutorial}.

\captionsetup[table]{skip=3pt}
\begin{table}[]
\centering
\resizebox{0.9\textwidth}{!}{%
\begin{tabular}{@{}lllllc@{}}
\toprule
 & CH-P & PeEn & MCM & \textbf{NGB} & \textbf{$\%\Delta$} \\ \midrule
Bondville, IL & 92.1 & 52.8 & 48.7 & \textbf{40.5} & -16.8\% \\
Boulder, CO & 91.3 & 61.6 & 51.6 & \textbf{45.9} & -11.0\% \\
Desert Rock, NV & 47.3 & 35.2 & 29.4 & \textbf{27.9} & -5.1\% \\
Fort Peck, MT & 77.0 & 46.3 & 39.8 & \textbf{34.8} & -12.6\% \\
Goodwin Creek, MS & 98.4 & 59.7 & 52.5 & \textbf{44.8} & -14.7\% \\
Penn State, PA & 98.1 & 60.0 & 53.0 & \textbf{46.0} & -13.2\% \\
Sioux Falls, SD & 86.8 & 47.8 & 41.0 & \textbf{37.9} & -7.6\% \\ \bottomrule
\end{tabular}%
 \quad\quad
\begin{tabular}{@{}llllll@{}}
\toprule
CH-P & GAU & NWP & NGB (\textit{+C})\\ \midrule
78.1 & 52.7 & \textbf{50.8} & 53.1 (52.9) \\
75.7 & 64.2 & 64.6 & \textbf{60.3} (\textbf{60.4}) \\
37.7 & 42.5 & 39.2 & \textbf{36.1} (\textbf{35.8}) \\
64.8 & 49.9 & 48.0 & \textbf{46.3} (\textbf{46.2}) \\
82.3 & 58.3 & \textbf{56.4} & 56.9 (56.6) \\
83.4 & \textbf{55.1} & 57.4 & 58.8 (58.1) \\
74.3 & 50.6 & \textbf{49.7} & 58.6 (56.6) \\\bottomrule
\end{tabular}%
}
\caption{\textbf{Per-Station CRPS Comparison}. The two tables show intra-hourly and hourly-precision results, respectively, with NGBoost denoted as NGB. \textit{+C} denotes NGBoost with CRUDE post-hoc calibration; other calibration methods underperformed CRUDE, and are included in the appendix. The non-NGBoost scores are taken from the best models in \cite{doubleday2020benchmark}, which are CH-P, PeEn, and MCM for intra-hourly and CH-P, GAU, and NWP for hourly forecasting. More details on the \cite{doubleday2020benchmark} models are provided in the appendix. \cite{doubleday2020benchmark} also evaluated their models on SURFRAD stations in 2018 during the daytime, with the same horizons and resolution.}
\vspace{-10px}
\label{ngboostvs}
\end{table}

\section{Discussion}
We evaluated several probabilistic models for solar irradiance forecasting combined with recent approaches for post-hoc calibration. We found that NGBoost, without post-hoc calibration, outperformed each of the baseline models across all of the stations at the intra-hourly resolution. Additionally, NGBoost achieved higher performance than two NWP model variants across three stations at the hourly-resolution. Our results suggest that NGBoost is an excellent baseline for probabilistic solar irradiance forecasting at both intra-hourly and hourly resolutions.

More sophisticated data and specialized priors could further improve our results, calling for more research on machine learning-based solar irradiance forecasting. The incorporation of satellite imagery with high temporal frequency\footnote{NOAA's GOES-16 and GOES-17 satellites launched in late 2016 and early 2018, respectively.} has the potential to improve intra-hourly uncertainty estimates \cite{yang2018history} by tracking the evolution of clouds which are the primary contributor to the variability of solar irradiance \cite{larson2020scope}. In addition, the priors used in this work to train the various models were Gaussian, but solar irradiance is never negative, and the clearness index is limited to 1; thus, a truncated normal distribution prior with fixed bounds could be more appropriate. Notably, NGBoost supports arbitrary priors as long as the derivative with respect to the output is calculable.

We believe that the development of probabilistic solar forecasting methods will help enable the level of renewable energy adoption that is necessary to phase out fossil fuel use. Furthermore, advancing probabilistic machine learning models and uncertainty estimation has implications to many problems related to climate change beyond solar forecasting \cite{rolnick2019tackling}. We hope our work helps to motivate further research in applied probabilistic machine learning research, which we believe is key to building technologies for mitigating climate change.

\begin{ack}
We would like to thank Cooper Elsworth, Kyle Story, and Rose Rustowicz from Descartes Labs for their help in the early directions of this work.
\end{ack}

{
\bibliography{bibliography}
\bibliographystyle{ieeetr}
}

\vfill
\pagebreak
\appendix
\section*{Appendix}
\setcounter{table}{0}
\setcounter{figure}{0}
\renewcommand{\thetable}{A\arabic{table}}
\renewcommand{\thefigure}{A\arabic{figure}}

\subsection*{Hyperparameters}
\textbf{Neural Network.} We train our neural network models using the Adam optimizer \cite{adamopt} with a learning rate of 1e-4 and for the variational neural network, a weight decay of 1e-2. We use three $1024$-wide hidden layers for the dropout-uncertainty neural network and three $256$-wide hidden layers for the variational neural network. 

\textbf{Gaussian Process.} We train our Gaussian process using the method proposed in \cite{hensman2015scalable}, using GPyTorch \cite{gardner2018gpytorch}, under a Gaussian probability distribution, with $1000$ batches of $1000$ points each, with early stopping after 10 batches without improvement.

\textbf{NGBoost.} For short term forecasting, we train NGBoost with the default hyperparameters and $2000$ estimators. For hourly forceasting, we also use a mini-batch fraction of $0.5$.

\textbf{Evaluation Details.} Each model was trained on each time horizon 10 times, and then calibrated and tested on random samples of $2,000$ points for each year. Only daytime data was evaluated (based on times when clear-sky irradiance was positive, as in \cite{doubleday2020benchmark}), but data from before sunrise was used for early-morning forecasts. On average, there are about $52,000$ daylight data points per station per year for intra-hourly forecasting, varying across stations and years by less than $\pm 1,000$.

\subsection*{Models from Doubleday \cite{doubleday2020benchmark}}
We compare to scores of the following models from \cite{doubleday2020benchmark}, though \cite{doubleday2020benchmark} includes further elaboration on each model:
\begin{itemize}[topsep=0pt,leftmargin=0.5cm]
    \item \textbf{CH-P.} CH-P corresponds to the complete-history persistence ensemble. It uses the historical distribution of clearness index values at a particular time of day at a particular station. 
    \item \textbf{PeEn.} PeEn corresponds to the persistence ensemble. Unlike the complete-history persistence ensemble, it uses only the most recent examples. For the intra-hourly case, this uses the clearness index values of the past two hours. 
    \item \textbf{MCM.} The Markov-chain mixture (MCM) model was first proposed in \cite{munkhammar2019probabilistic}. MCM attempts to model implicit states and their associated transition probabilities.
    \item \textbf{NWP.} The numerical weather prediction (NWP) model used in \cite{doubleday2020benchmark} was an ensemble of the 51 predictions by the European Centre for Medium-Range Weather Forecasts (ECMWF) predictions (the control forecast and the 50 perturbed forecasts).
    \item \textbf{GAU.} The Gaussian error distribution for the hourly forecasts exclusively uses the ECMWF unperturbed control forecast, alongside a standard deviation derived from the distribution of errors for the same time of day over the course of a year. It uses a double-truncated Gaussian distribution to exclude negative or very positive clearness index values. 
\end{itemize}

\subsection*{Calibration Curves}

We include some example calibration curves to help visualize the impact of post-hoc calibration on different models.

\begin{figure}[h]
  \centering
  \includegraphics[width=0.45\textwidth]{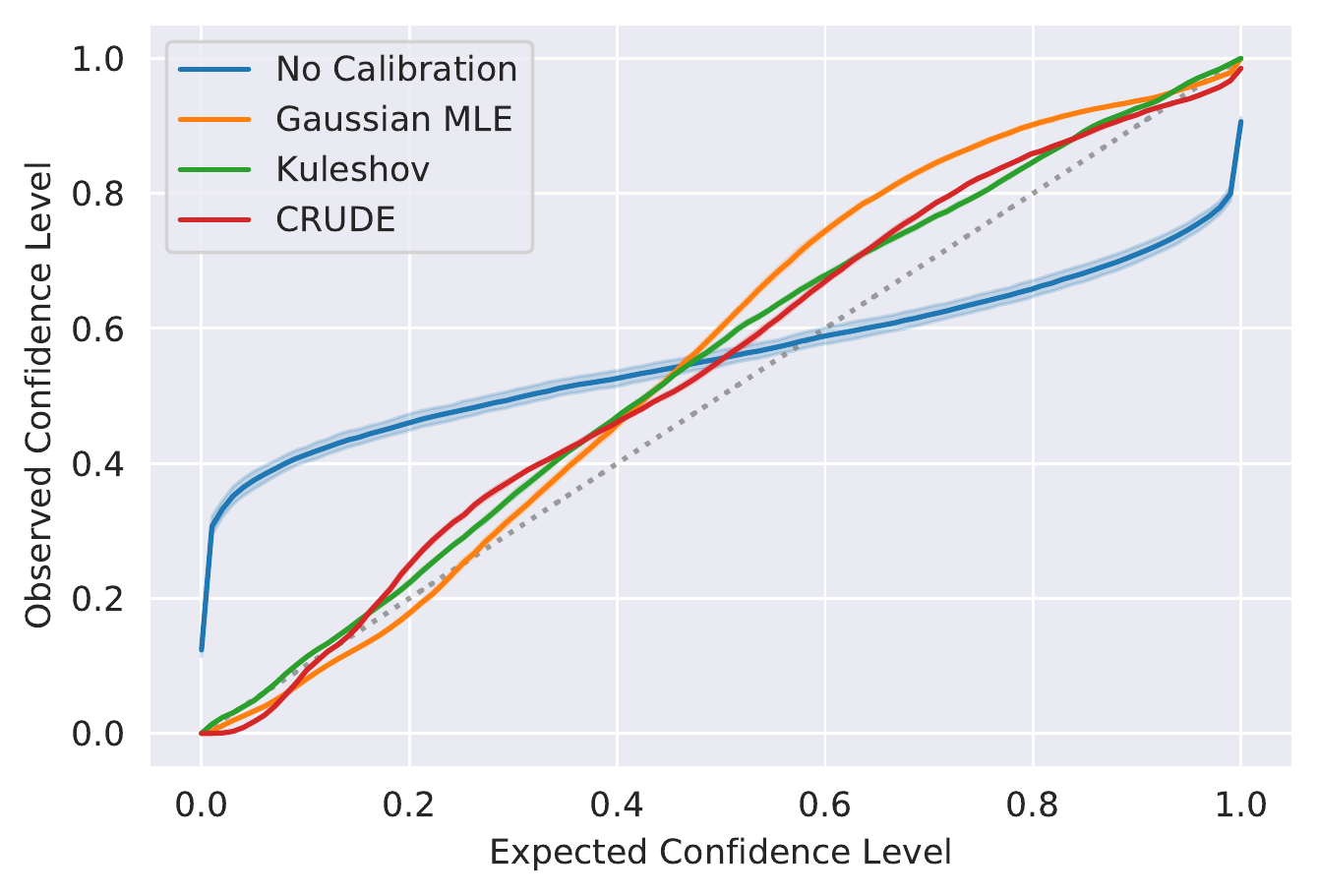} 
  \includegraphics[width=0.45\textwidth]{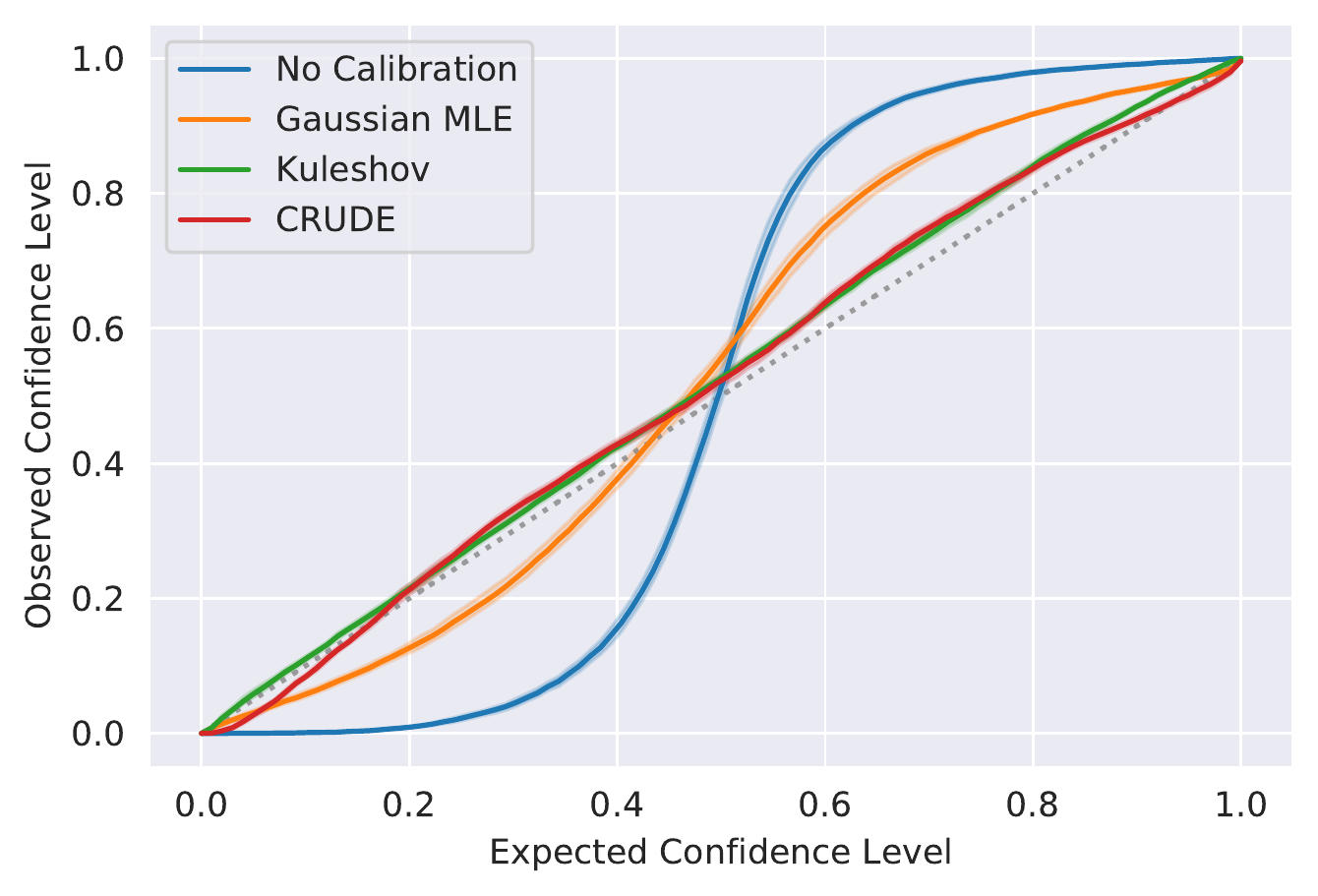}
  \includegraphics[width=0.45\textwidth]{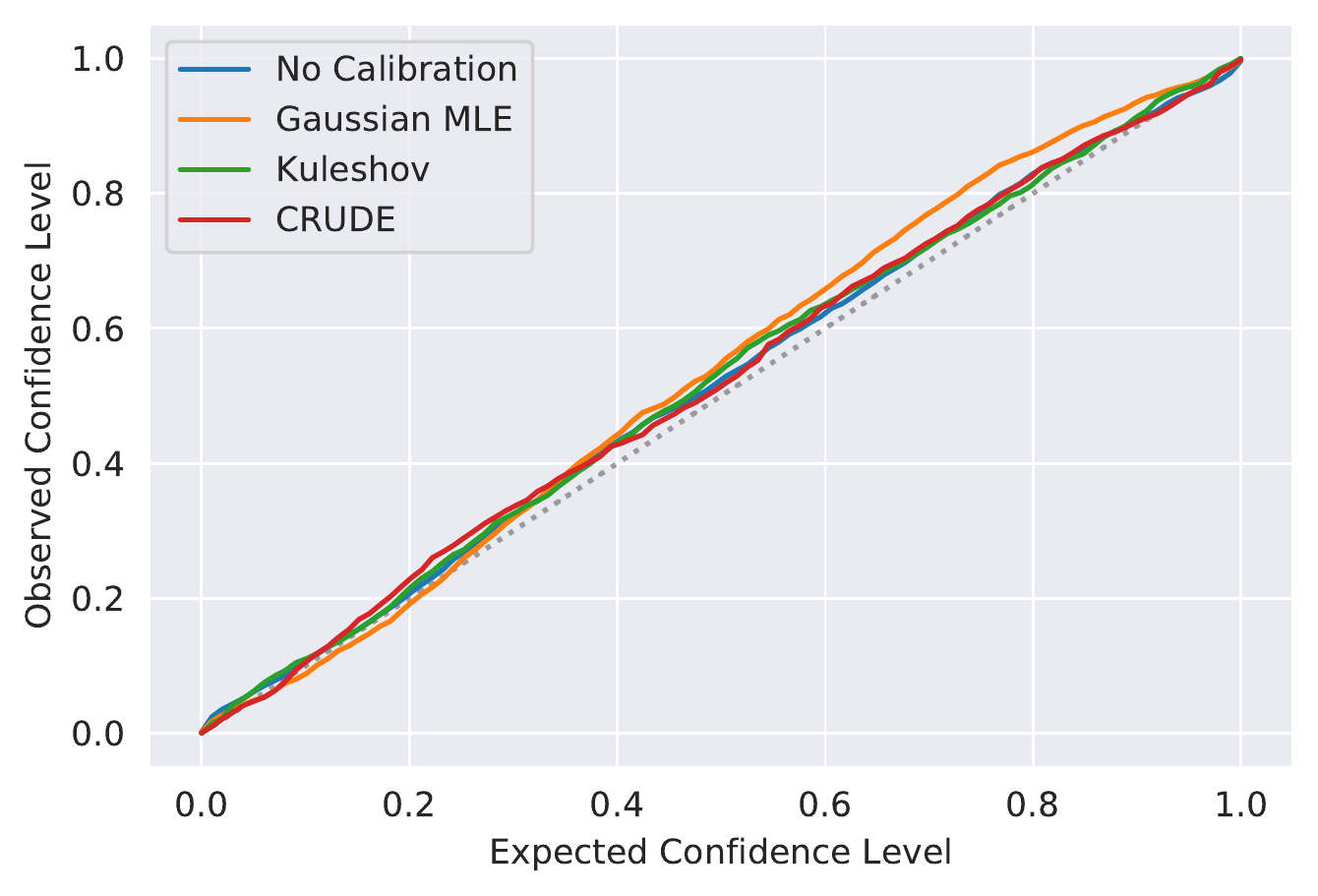}
  \includegraphics[width=0.45\textwidth]{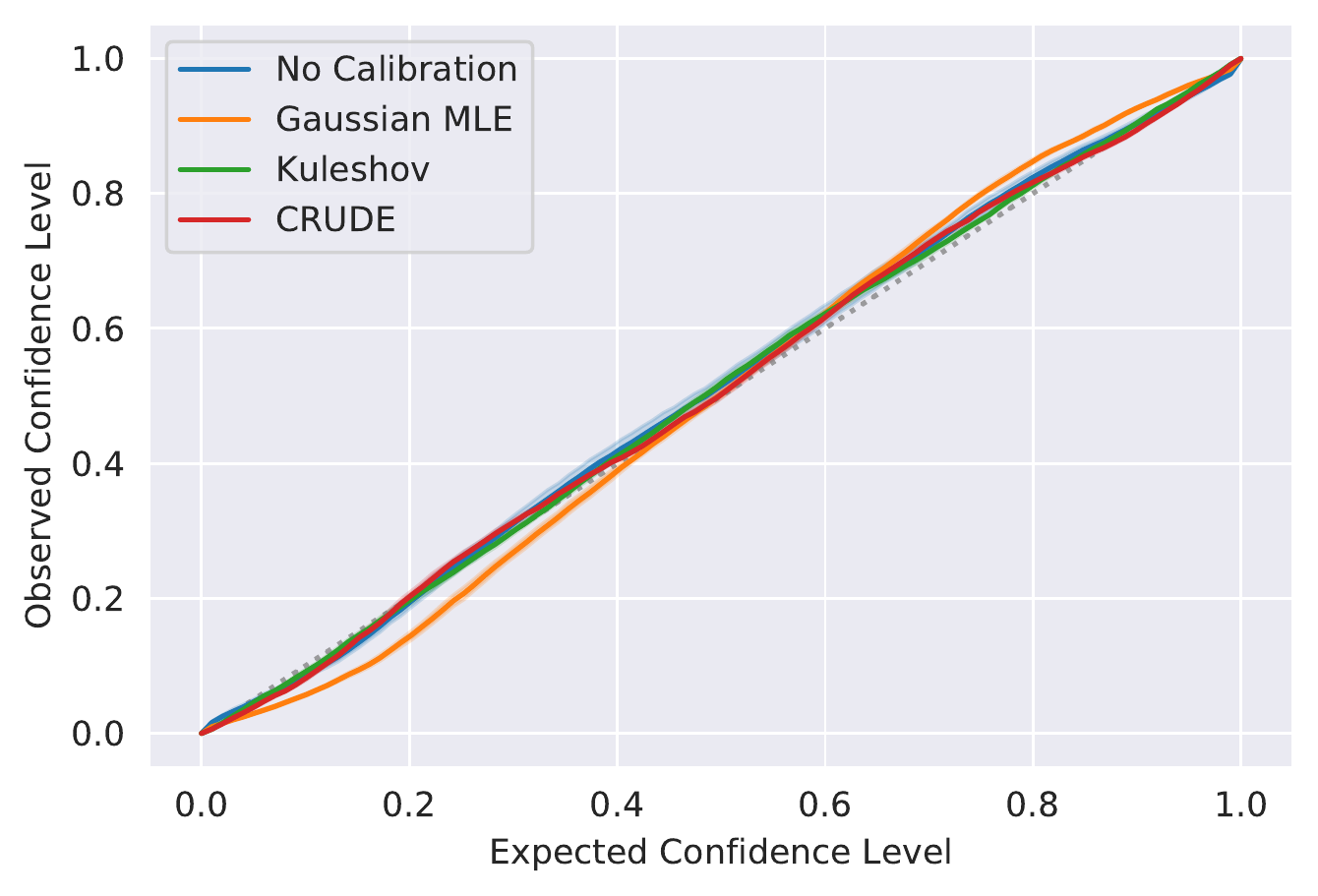}
  \caption{Calibration curves for the dropout neural network (top left), the Gaussian process (top right), NGBoost (bottom left), and the variational neural network (bottom right) on the Penn State, PA SURFRAD station with a 30 minute horizon.}
  \label{calcurves}
\end{figure}

\vfill
\pagebreak
\subsection*{Hourly Forecasting with NGBoost }

\captionsetup[table]{skip=3pt}
\begin{table}[h]
\centering
\resizebox{0.5\textwidth}{!}{%
\begin{tabular}{@{}lcccc@{}}
\toprule
 & \textit{None} & MLE & CRUDE & Kuleshov \\ \midrule
Bondville, IL & 53.1 & 53.8 & \textbf{52.9} & 53.0 \\
Boulder, CO & \textbf{60.3} & 68.0 & 60.4 & 61.7\\
Desert Rock, NV & 36.1 & 37.3 & \textbf{35.8} & 37.0 \\
Fort Peck, MT & 46.3 & 46.3 & \textbf{46.2} & 46.9\\
Goodwin Creek, MS & 56.9 & 56.8 & \textbf{56.6} & 57.8 \\
Penn State, PA & 58.8 & \textbf{58.1} & \textbf{58.1} & 58.3 \\
Sioux Falls, SD & 58.6 & 57.8 & \textbf{56.6} & 57.7 \\ \bottomrule
\end{tabular}%
}
\caption{\textbf{Hourly NBoost forecast CRPS by station}. A comparison of the CRPS of each calibration method applied to NGBoost on each station.}
\vspace{-5px}
\end{table}

As mentioned in the main text, CRUDE consistently outperformed the other calibration methods. Curiously, while there was no significant positive impact by calibration on NGBoost for intra-hourly forecasting, there was a notable improvement in the hourly forecasting case. It is possible that there is less yearly variation in this longer timescale, and thus calibration is more effective. This suggests that training and calibrating over a period of more years would improve performance. Note that the difference between the intra-hourly curve and the hourly curve at a horizon of one hour comes from aggregation: the intra-hourly metric evalautes the 5-minute average CRPS, while the hourly metric predicts the one-hour average CRPS.

\begin{figure}[h]
  \centering
  \includegraphics[width=0.35\textwidth]{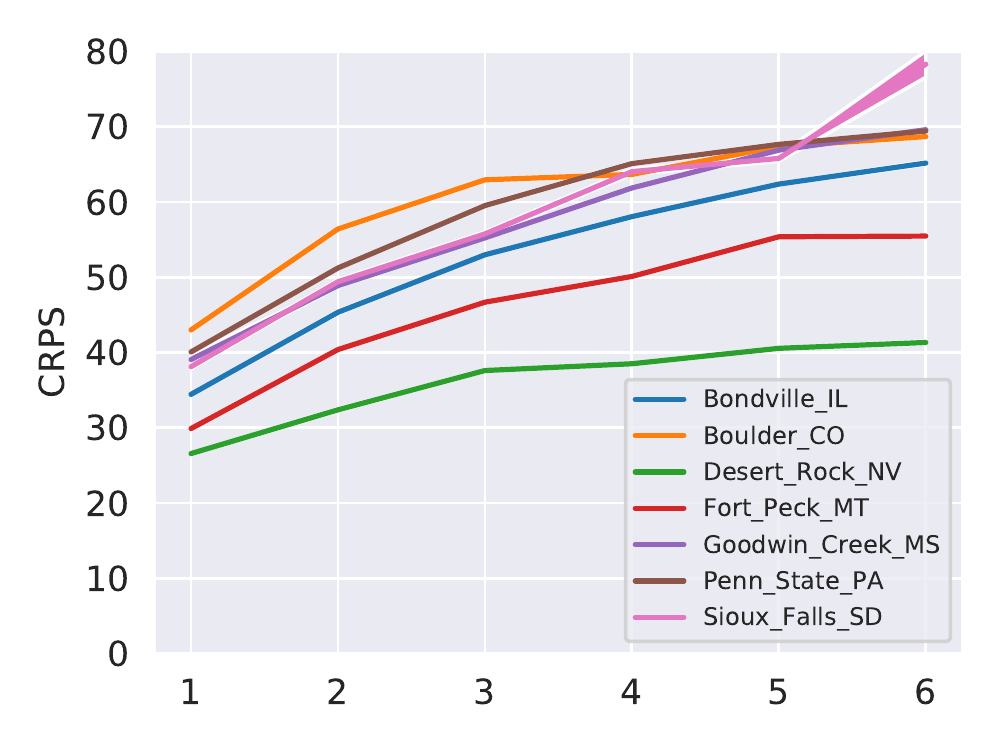}
  \includegraphics[width=0.35\textwidth]{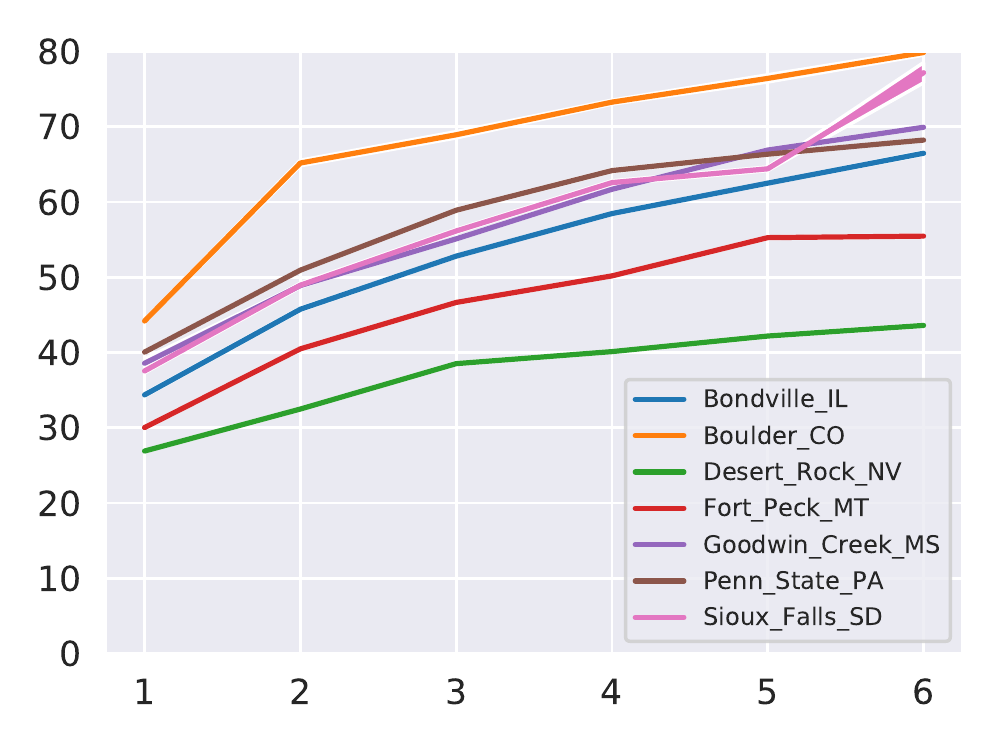}
  \includegraphics[width=0.35\textwidth]{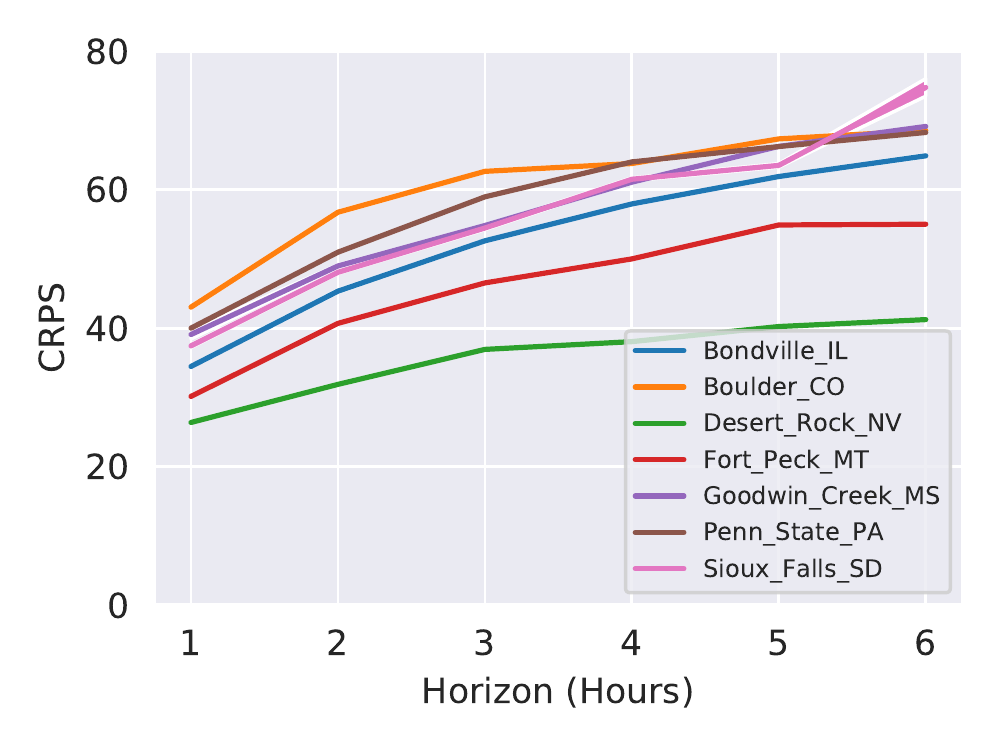} 
  \includegraphics[width=0.35\textwidth]{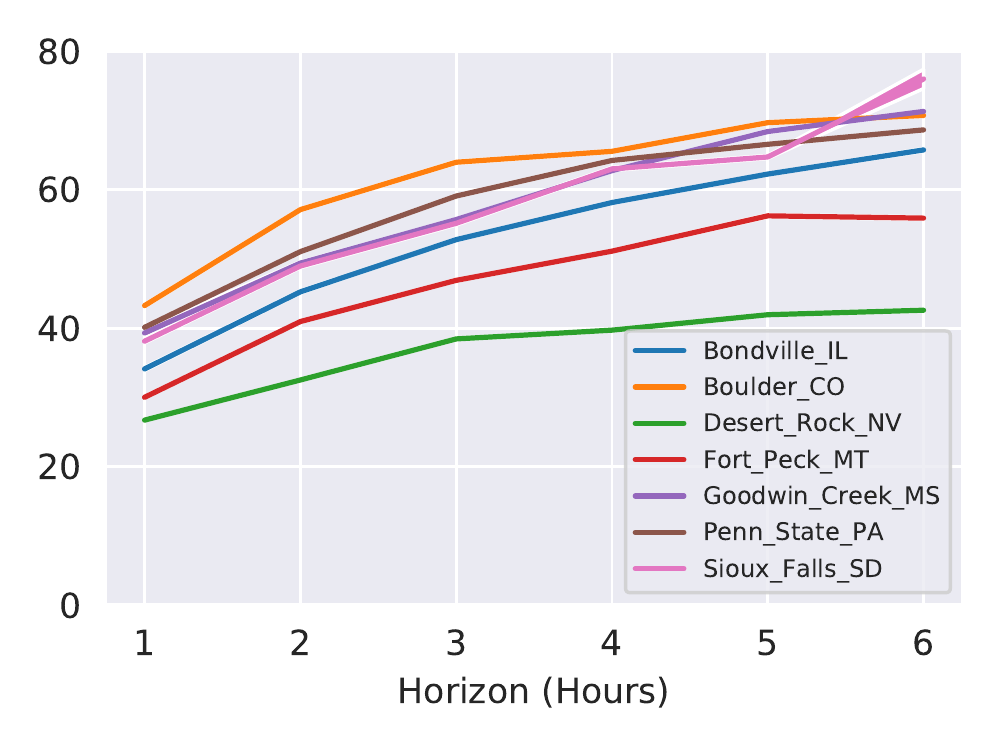}
  \caption{Performance of various hourly NGBoost calibration methods over all stations. These plots also show the variance in CRPS scores over each of a model's 10 evaluations at a station and horizon - notably this $\pm 1 \sigma$ is generally not visible for NGBoost.}
\end{figure}

\vfill
\pagebreak
\subsection*{Calibration and Sharpness}

\textbf{Calibration}
\begin{table}[h]
\hspace{-10px}
\resizebox{1.05 \textwidth}{!}{%
\begin{tabular}{@{}lrrrrrrrrrrrrrrrr@{}}
\toprule
                                        & \multicolumn{4}{c}{\textbf{Gaussian Process}}                                                           & \multicolumn{4}{c}{\textbf{Dropout Neural Network}}                                                     & \multicolumn{4}{c}{\textbf{Variational Neural Net}}                                                     & \multicolumn{4}{c}{\textbf{NGBoost}}                                                                  \\ \midrule
\multicolumn{1}{l|}{\textit{}}          & \multicolumn{1}{l}{None} & \multicolumn{1}{l}{MLE} & \multicolumn{1}{l}{C} & \multicolumn{1}{l|}{Kul.}  & \multicolumn{1}{l}{None} & \multicolumn{1}{l}{MLE} & \multicolumn{1}{l}{C} & \multicolumn{1}{l|}{Kul.}  & \multicolumn{1}{l}{None} & \multicolumn{1}{l}{MLE} & \multicolumn{1}{l}{C} & \multicolumn{1}{l|}{Kul.}  & \multicolumn{1}{l}{None} & \multicolumn{1}{l}{MLE} & \multicolumn{1}{l}{C} & \multicolumn{1}{l}{Kul.} \\
\multicolumn{1}{l|}{Bondville, IL}     & 0.19                     & 0.08                    & 0.01                  & \multicolumn{1}{r|}{0.03} & 0.15                     & 0.07                    & 0.02                  & \multicolumn{1}{r|}{0.04} & 0.04                     & 0.05                    & 0.02                  & \multicolumn{1}{r|}{0.03} & 0.04                     & 0.05                    & 0.02                  & 0.03                     \\
\multicolumn{1}{l|}{Boulder, CO}       & 0.18                     & 0.09                    & 0.03                  & \multicolumn{1}{r|}{0.04} & 0.14                     & 0.07                    & 0.03                  & \multicolumn{1}{r|}{0.05} & 0.05                     & 0.07                    & 0.04                  & \multicolumn{1}{r|}{0.05} & 0.04                     & 0.06                    & 0.03                  & 0.03                     \\
\multicolumn{1}{l|}{Desert Rock,NV}    & 0.21                     & 0.13                    & 0.04                  & \multicolumn{1}{r|}{0.04} & 0.10                     & 0.11                    & 0.03                  & \multicolumn{1}{r|}{0.04} & 0.10                     & 0.12                    & 0.04                  & \multicolumn{1}{r|}{0.04} & 0.05                     & 0.11                    & 0.03                  & 0.04                     \\
\multicolumn{1}{l|}{Fort Peck, MT}     & 0.19                     & 0.10                    & 0.02                  & \multicolumn{1}{r|}{0.03} & 0.13                     & 0.09                    & 0.03                  & \multicolumn{1}{r|}{0.04} & 0.05                     & 0.13                    & 0.02                  & \multicolumn{1}{r|}{0.02} & 0.03                     & 0.05                    & 0.02                  & 0.03                     \\
\multicolumn{1}{l|}{Goodwin Creek, MS} & 0.19                     & 0.08                    & 0.02                  & \multicolumn{1}{r|}{0.03} & 0.15                     & 0.07                    & 0.02                  & \multicolumn{1}{r|}{0.03} & 0.05                     & 0.05                    & 0.03                  & \multicolumn{1}{r|}{0.03} & 0.03                     & 0.05                    & 0.03                  & 0.04                     \\
\multicolumn{1}{l|}{Penn State, PA}    & 0.19                     & 0.08                    & 0.02                  & \multicolumn{1}{r|}{0.03} & 0.17                     & 0.08                    & 0.04                  & \multicolumn{1}{r|}{0.04} & 0.04                     & 0.05                    & 0.03                  & \multicolumn{1}{r|}{0.03} & 0.02                     & 0.05                    & 0.03                  & 0.03                     \\
\multicolumn{1}{l|}{Sioux Falls, SD}   & 0.19                     & 0.08                    & 0.04                  & \multicolumn{1}{r|}{0.05} & 0.19                     & 0.10                    & 0.09                  & \multicolumn{1}{r|}{0.09} & 0.12                     & 0.07                    & 0.10                  & \multicolumn{1}{r|}{0.10} & 0.07                     & 0.08                    & 0.07                  & 0.08                     \\ \bottomrule
\end{tabular}%
}
\caption{\textbf{Per-Station Calibration Comparison}. We include a comparison of the stations and their calibration scores. Lower scores correspond to better calibration. The abbreviations correspond to those used in Table~\ref{stations-models}. }
\vspace{-20px}

\end{table}

\textbf{Sharpness}
\begin{table}[h]
\hspace{-10px}
\resizebox{1.05 \textwidth}{!}{%
\begin{tabular}{@{}lrrrrrrrrrrrrrrrr@{}}
\toprule
                                        & \multicolumn{4}{c}{\textbf{Gaussian Process}}                                                              & \multicolumn{4}{c}{\textbf{Dropout Neural Network}}                                                        & \multicolumn{4}{c}{\textbf{Variational Neural Net}}                                                        & \multicolumn{4}{c}{\textbf{NGBoost}}                                                                      \\ \midrule
\multicolumn{1}{l|}{\textit{}}          & \multicolumn{1}{l}{None} & \multicolumn{1}{l}{MLE} & \multicolumn{1}{l}{C} & \multicolumn{1}{l|}{Kul.} & \multicolumn{1}{l}{None} & \multicolumn{1}{l}{MLE} & \multicolumn{1}{l}{C} & \multicolumn{1}{l|}{Kul.} & \multicolumn{1}{l}{None} & \multicolumn{1}{l}{MLE} & \multicolumn{1}{l}{C} & \multicolumn{1}{l|}{Kul.} & \multicolumn{1}{l}{None} & \multicolumn{1}{l}{MLE} & \multicolumn{1}{l}{C} & \multicolumn{1}{l}{Kul.} \\
\multicolumn{1}{l|}{Bondville, IL}     & 0.99                     & 0.36                    & 0.28                  & \multicolumn{1}{r|}{0.35} & 0.06                     & 0.25                    & 0.23                  & \multicolumn{1}{r|}{0.25} & 0.21                     & 0.24                    & 0.22                  & \multicolumn{1}{r|}{0.24}  & 0.20                     & 0.23                    & 0.21                  & 0.22                     \\
\multicolumn{1}{l|}{Boulder, CO}       & 1.02                     & 0.39                    & 0.31                  & \multicolumn{1}{r|}{0.39} & 0.06                     & 0.27                    & 0.25                  & \multicolumn{1}{r|}{0.26} & 0.25                     & 0.30                    & 0.25                  & \multicolumn{1}{r|}{0.30}  & 0.21                     & 0.25                    & 0.22                  & 0.25                     \\
\multicolumn{1}{l|}{Desert Rock,NV}    & 0.91                     & 0.31                    & 0.25                  & \multicolumn{1}{r|}{0.30} & 0.07                     & 0.21                    & 0.20                  & \multicolumn{1}{r|}{0.21} & 0.22                     & 0.26                    & 0.21                  & \multicolumn{1}{r|}{0.26}  & 0.16                     & 0.27                    & 0.19                  & 0.27                     \\
\multicolumn{1}{l|}{Fort Peck, MT}     & 1.07                     & 0.42                    & 0.29                  & \multicolumn{1}{r|}{0.39} & 0.07                     & 0.27                    & 0.26                  & \multicolumn{1}{r|}{0.26} & 0.26                     & 0.60                    & 0.19                  & \multicolumn{1}{r|}{0.24} & 0.19                     & 0.24                    & 0.21                  & 0.24                     \\
\multicolumn{1}{l|}{Goodwin Creek, MS} & 1.10                     & 0.38                    & 0.30                  & \multicolumn{1}{r|}{0.37} & 0.06                     & 0.26                    & 0.25                  & \multicolumn{1}{r|}{0.26} & 0.23                     & 0.25                    & 0.23                  & \multicolumn{1}{r|}{0.24}  & 0.20                     & 0.22                    & 0.21                  & 0.22                     \\
\multicolumn{1}{l|}{Penn State, PA}    & 1.07                     & 0.40                    & 0.31                  & \multicolumn{1}{r|}{0.40} & 0.06                     & 0.27                    & 0.25                  & \multicolumn{1}{r|}{0.27} & 0.23                     & 0.27                    & 0.24                  & \multicolumn{1}{r|}{0.27}  & 0.20                     & 0.24                    & 0.22                  & 0.23                     \\
\multicolumn{1}{l|}{Sioux Falls, SD}   & 1.08                     & 0.39                    & 0.30                  & \multicolumn{1}{r|}{0.38} & 0.07                     & 0.26                    & 0.26                  & \multicolumn{1}{r|}{0.26} & 0.20                     & 0.30                    & 0.23                  & \multicolumn{1}{r|}{0.28}  & 0.19                     & 0.27                    & 0.23                  & 0.27                     \\ \bottomrule
\end{tabular}%
}
\caption{\textbf{Per-Station Sharpness Comparison}. We include a comparison of the stations and their sharpness scores. We report sharpness in terms of clearness index. Note that sharpness is only meaningful under a well-calibrated model. Lower scores correspond to sharper models. The abbreviations also correspond to those used in Table~\ref{stations-models}.}
\vspace{-12px}
\end{table}

As discussed in the main text, we analyzed and compared sharpness and calibration metrics, based on \cite{kuleshov2018accurate, crude}. In general, NGBoost is both slightly better calibrated and sharper than the variational neural network model, which corresponds to their CRPS performance. Generally, with post-hoc calibration, the variational neural network and NGBoost were sharper than the Gaussian process and dropout-uncertainty based neural network. We also see that the Gaussian process model is underconfident without post-hoc calibration, while the dropout-uncertainty neural network is overconfident. 
\end{document}